\documentclass[11pt,letterpaper]{article}
\usepackage{emnlp2016}
\usepackage{times}
\usepackage{latexsym}

\usepackage{amsmath}
\usepackage{amssymb}
\usepackage{amsopn}
\usepackage{subcaption}
\usepackage{CJKutf8}
\usepackage{graphicx}
\usepackage{multirow}
\usepackage{color}
\usepackage{algorithm}  
\usepackage{algorithmic}  
\usepackage[skip=2pt,belowskip=-12pt,aboveskip=2pt,font=small]{caption}

\newcommand{\chinese}[1]{{\begin{CJK*}{UTF8}{gkai}#1\end{CJK*}}}
\definecolor{zb_red}{RGB}{200, 0, 0}

\usepackage{enumitem}
\setenumerate[1]{itemsep=0.30pt,partopsep=0pt,parsep=\parskip,topsep=5pt}
\setitemize[1]{itemsep=0.30pt,partopsep=00pt,parsep=\parskip,topsep=5pt}
\setdescription{itemsep=0.30pt,partopsep=0pt,parsep=\parskip,topsep=5pt} 

\emnlpfinalcopy

\def\emnlppaperid{***}

\title{Variational Neural Machine Translation}

\author{Biao Zhang$^{1,2}$, Deyi Xiong$^{1}$\thanks{~~Corresponding author}, Jinsong Su$^{2}$, Hong Duan$^{2}$ and Min Zhang$^{1}$\\
	Provincial Key Laboratory for Computer Information Processing Technology \\
	Soochow University, Suzhou, China 215006$^{1}$ \\
	Xiamen University, Xiamen, China 361005$^{2}$ \\
	{\tt zb@stu.xmu.edu.cn, \{jssu,hduan\}@xmu.edu.cn} \\
	{\tt \{dyxiong, minzhang\}@suda.edu.cn} \\
}

\date{}

\begin{document}

\maketitle

\begin{abstract}
Models of neural machine translation are often from a discriminative family of encoder-decoders that learn a conditional distribution of a target sentence given a source sentence. In this paper, we propose a variational model to learn this conditional distribution for neural machine translation: a variational encoder-decoder model that can be trained end-to-end. Different from the vanilla encoder-decoder model that generates target translations from hidden representations of source sentences alone, the variational model introduces a {\it continuous latent variable} to explicitly model underlying semantics of source sentences and to guide the generation of target translations. 
In order to perform efficient posterior inference and large-scale training, 
we build a {\it neural posterior approximator} conditioned on both the source and the target sides,
and equip it with a reparameterization technique to estimate the variational lower bound. 
Experiments on both Chinese-English and English-German translation tasks show that the proposed variational neural machine translation achieves significant improvements over the vanilla neural machine translation baselines. 
\end{abstract}

\section{Introduction}

Neural machine translation (NMT) is an emerging translation paradigm that builds on a single and unified end-to-end neural network, instead of using a variety of sub-models tuned in a long training pipeline. It requires a much smaller memory than phrase- or syntax-based statistical machine translation (SMT) that typically has a huge phrase/rule table. Due to these advantages over traditional SMT system,  NMT has recently attracted growing interests from both deep learning and machine translation community~\cite{kalchbrenner-blunsom:2013:EMNLP,cho-EtAl:2014:EMNLP2014,DBLP:journals/corr/SutskeverVL14,DBLP:journals/corr/BahdanauCB14,luong-pham-manning:2015:EMNLP,luong-EtAl:2015:ACL-IJCNLP,2015arXiv151202433S,2015arXiv150606442M,DBLP:journals/corr/TuLLLL16}.

Current NMT models mainly take a discriminative {\it encoder}-{\it decoder} framework, where a {\it neural encoder} transforms source sentence $\mathbf{x}$ into distributed representations, and a {\it neural decoder} generates the corresponding target sentence $\mathbf{y}$ according to these representations\footnote{In this paper, we use bold symbols to denote variables, and plain symbols to denote their values. Without specific statement, all variables are multivariate.}~\cite{cho-EtAl:2014:EMNLP2014,DBLP:journals/corr/SutskeverVL14,DBLP:journals/corr/BahdanauCB14}. Typically, the underlying semantic representations of source and target sentences are learned in an implicit way in this framework, which heavily relies on the attention mechanism~\cite{DBLP:journals/corr/BahdanauCB14} to identify semantic alignments between source and target words. Due to potential errors in these alignments, the attention-based context vector may be insufficient to capture the entire meaning of a source sentence, hence  resulting in undesirable translation phenomena~\cite{DBLP:journals/corr/TuLLLL16}.



Unlike the vanilla encoder-decoder framework, we model underlying semantics of bilingual sentence pairs explicitly. We assume that there exists a continuous latent variable $\mathbf{z}$ from this underlying semantic space. And this variable, together with $\mathbf{x}$, guides the translation process, i.e. $p(\mathbf{y}|\mathbf{z}, \mathbf{x})$. With this assumption, the original conditional probability evolves into the following formulation:
\begin{equation}\label{vnmt_form}
p(\mathbf{y}|\mathbf{x})  = \int_{z} p(\mathbf{y},{z}|\mathbf{x})d_{z} = \int_{z} p(\mathbf{y}|{z},\mathbf{x}) p({z}|\mathbf{x})d_{z}
\end{equation}
This brings in the benefits that the latent variable $\mathbf{z}$ can serve as a global semantic signal that is complementary to the attention-based context vector for generating good translations when the model learns undesirable attentions. However, although this latent variable enables us to explicitly model underlying semantics of translation pairs, the incorporation of it into the above probabilistic model has two challenges: 1) the posterior inference in this model is intractable; 2) large-scale training, which lays the ground for the data-driven NMT, is accordingly problematic. 

\begin{figure}[t]
\centering
\includegraphics[scale=0.70]{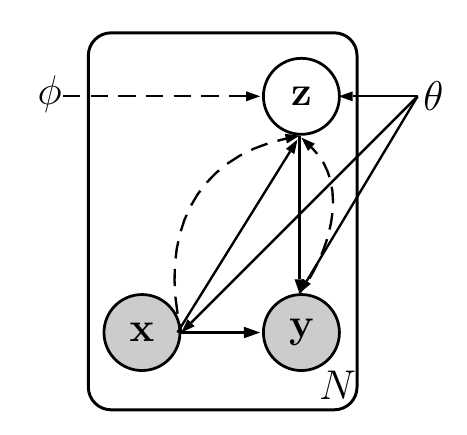}
\caption{\label{graph_model} Illustration of VNMT as a directed graph. We use solid lines to denote the generative model $p_\theta(\mathbf{z}|\mathbf{x})p_\theta(\mathbf{y}|\mathbf{z},\mathbf{x})$, and dashed lines to denote the variational approximation $q_\phi(\mathbf{z}|\mathbf{x})$ to the intractable posterior $p(\mathbf{z}|\mathbf{x}, \mathbf{y})$. Both variational parameters $\phi$ and generative model parameters $\theta$ are learned jointly.}
\end{figure}

\begin{figure*}[t]
\centering
\includegraphics[scale=0.70]{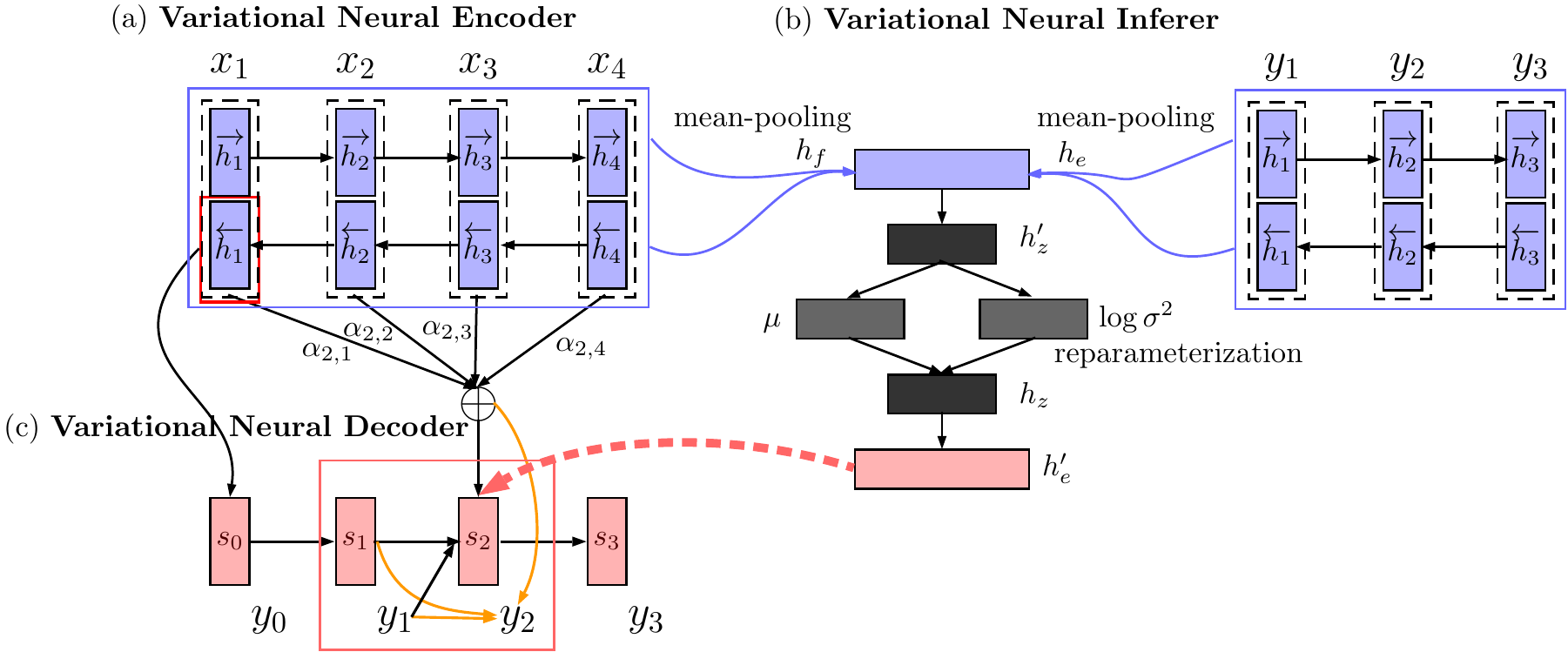}
\caption{\label{detail_model} Neural architecture of VNMT. We use blue, gray and red color to indicate the encoder-related ($\mathbf{x},\mathbf{y}$), underlying semantic ($\mathbf{z}$) and decoder-related ($\mathbf{y}$) representation respectively. The yellow lines show the flow of information employed for target word prediction. The dashed red line highlights the incorporation of latent variable $\mathbf{z}$ into target prediction. $f$ and $e$ represent the source and target language respectively.
}
\end{figure*}

In order to address these issues, we propose a variational encoder-decoder model to neural machine translation (VNMT), motivated by the recent success of variational neural models~\cite{DBLP:conf/icml/RezendeMW14,kingma2014autoencoding}. Figure \ref{graph_model} illustrates the graphic representation of VNMT. 
As deep neural networks are capable of learning highly non-linear functions, we employ them to fit the latent-variable-related distributions, i.e. the prior and posterior, to make the inference tractable. The former is modeled to be conditioned on the source side alone $p_\theta(\mathbf{z}|\mathbf{x})$, because the source and target part of a sentence pair usually share the same semantics so that the source sentence should contain the prior information for inducing the underlying semantics. The latter, instead, is approximated from all observed variables $q_\phi(\mathbf{z}|\mathbf{x}, \mathbf{y})$, i.e. both the source and the target sides.
In order to efficiently train parameters, we apply a reparameterization technique~\cite{DBLP:conf/icml/RezendeMW14,kingma2014autoencoding} on the variational lower bound. This enables us to use standard stochastic gradient optimization for training the proposed model. Specifically, there are three essential components in VNMT (The detailed architecture is illustrated in Figure \ref{detail_model}):
\begin{itemize}
\item A {\it variational neural encoder} transforms source/target sentence into distributed representations, which is the same as the encoder of NMT~\cite{DBLP:journals/corr/BahdanauCB14} (see section \ref{var_neu_encoder}).
\item A {\it variational neural inferer} infers the representation of $\mathbf{z}$ according to the learned source representations (i.e. $p_\theta($$\mathbf{z}$$|$$\mathbf{x})$) together with the target ones (i.e. $q_\phi($$\mathbf{z}$$|$$\mathbf{x},\mathbf{y})$), where the reparameterization technique is employed (see section \ref{var_neu_approx}).
\item And a {\it variational neural decoder} integrates the latent representation of $\mathbf{z}$ to guide the generation of target sentence (i.e. $p(\mathbf{y}|\mathbf{z},\mathbf{x})$) together with the attention mechanism (see section \ref{var_neu_decoder}).
\end{itemize}

Augmented with the posterior approximation and reparameterization, our VNMT can still be trained end-to-end. This makes our model not only efficient in translation, but also simple in implementation. To train our model, we employ the conventional maximum likelihood estimation. Experiments on both Chinese-English and English-German translation tasks show that VNMT achieves significant improvements over several strong baselines. 


\section{Background: Variational Autoencoder}

This section briefly reviews the variational autoencoder (VAE)~\cite{kingma2014autoencoding,DBLP:conf/icml/RezendeMW14}. 
Given an observed variable $\mathbf{x}$, VAE introduces a continuous latent variable $\mathbf{z}$, and assumes that $\mathbf{x}$ is generated from $\mathbf{z}$, i.e.,
\begin{equation}\label{vae_joint}
p_\theta(\mathbf{x}, \mathbf{z}) = p_\theta(\mathbf{x}|\mathbf{z})p_\theta(\mathbf{z})
\end{equation}
where $\theta$ denotes the parameters of the model. $p_\theta(\mathbf{z})$ is the prior, e.g, a simple Gaussian distribution. $p_\theta(\mathbf{x}|\mathbf{z})$ is the conditional distribution that models the generation procedure, typically estimated via a deep non-linear neural network.

Similar to our model, the integration of $\mathbf{z}$ in Eq. (\ref{vae_joint}) imposes challenges on the posterior inference as well as large-scale learning. To tackle these problems, VAE adopts two techniques: {\it neural approximation} and {\it reparameterization}.

{\it Neural Approximation} employs deep neural networks to approximate the posterior inference model $q_\phi(\mathbf{z}|\mathbf{x})$, where $\phi$ denotes the variational parameters. For the posterior approximation, VAE regards $q_\phi(\mathbf{z}|\mathbf{x})$ as a diagonal Gaussian $\mathcal{N}(\mathbf{\mu}, \text{diag}(\mathbf{\sigma^2}))$, and parameterizes its mean $\mathbf{\mu}$ and variance $\mathbf{\sigma^2}$ with deep neural networks. 

{\it Reparameterization} reparameterizes $\mathbf{z}$ as a function of $\mathbf{\mu}$ and $\mathbf{\sigma}$, rather than using the standard sampling method. In practice, VAE leverages the ``location-scale'' property of Gaussian distribution, and uses the following reparameterization:
\begin{equation}\label{vae_re_param}
\tilde{{z}} = \mathbf{\mu} + \mathbf{\sigma}\odot \mathbf{\epsilon}
\end{equation}
where $\mathbf{\epsilon}$ is a standard Gaussian variable that plays a role of introducing noises, and $\odot$ denotes an element-wise product. 

With these two techniques, VAE tightly incorporates both the generative model $p_\theta(\mathbf{x}|\mathbf{z})$ and the posterior inference model $q_\phi(\mathbf{z}|\mathbf{x})$ into an end-to-end neural network. This facilitates its optimization since we can apply the standard backpropagation to compute the gradient of the following variational lower bound:
\begin{equation}\label{vae_vlb}
\begin{split}
\mathcal{L}_{\text{\tiny VAE}}(\theta,\phi;\mathbf{x}) = &-\text{KL}(q_\phi(\mathbf{z}|\mathbf{x})||p_\theta(\mathbf{z})) \\
+ \mathbb{E}_{q_\phi(\mathbf{z}|\mathbf{x})}&[\log p_\theta(\mathbf{x}|\mathbf{z})] \leq \log p_\theta(\mathbf{x})
\end{split}
\end{equation}
$\text{KL}(Q||P)$ is the Kullback-Leibler divergence between $Q$ and $P$. Intuitively, VAE can be considered as a regularized version of the standard autoencoder. It makes use of the latent variable $\mathbf{z}$ to capture the variations $\mathbf{\epsilon}$ in the observed variable $\mathbf{x}$.


\section{Variational Neural Machine Translation}

Different from previous work, we introduce a latent variable $\mathbf{z}$ to model the underlying semantic space as a global signal for translation. Formally, given the definition in Eq. (\ref{vnmt_form}) and Eq. (\ref{vae_vlb}), the variational lower bound of VNMT can be formulated as follows:
\begin{align} \label{vnmt_vlb_org}
\mathcal{L}_{\text{\tiny VNMT}}(\theta,\phi;\mathbf{x},\mathbf{y}) = -\text{KL}&(q_\phi(\mathbf{z}|\mathbf{x},\mathbf{y})||p_\theta(\mathbf{z}|\mathbf{x})) \notag \\
+ \mathbb{E}_{q_\phi(\mathbf{z}|\mathbf{x},\mathbf{y})}&[\log p_\theta(\mathbf{y}|\mathbf{z},\mathbf{x})]
\end{align}
where $p_\theta(\mathbf{z}|\mathbf{x})$ is our prior model, $q_\phi(\mathbf{z}|\mathbf{x}, \mathbf{y})$ is our posterior approximator, and $p_\theta(\mathbf{y}|\mathbf{z},\mathbf{x})$ is the decoder with the guidance from $\mathbf{z}$. Based on this formulation, VNMT can be decomposed into three components, each of which is modeled by a neural network: a {\it variational neural inferer} that models $p_\theta(\mathbf{z}|\mathbf{x})$ and $q_\phi(\mathbf{z}|\mathbf{x}, \mathbf{y})$ (see part (b) in Figure \ref{detail_model}), a {\it variational neural decoder} that models $p_\theta(\mathbf{y}|\mathbf{z},\mathbf{x})$ (see part (c) in Figure \ref{detail_model}), and a {\it variational neural encoder} that provides distributed representations of a source/target sentence for the above two modules (see part (a) in Figure \ref{detail_model}). Following the information flow illustrated in Figure \ref{detail_model}, we describe part (a), (b) and (c) successively.


\subsection{Variational Neural Encoder}\label{var_neu_encoder}

As shown in Figure \ref{detail_model} (a), the variational neural encoder aims at encoding an input sequence $(w_1,$ $w_2,$ $\ldots,$ $w_{T})$ into continuous vectors. In this paper, we adopt the encoder architecture proposed by Bahdanau et al.~\shortcite{DBLP:journals/corr/BahdanauCB14}, which is a bidirectional RNN with a forward and backward RNN. The forward RNN reads the sequence from left to right while the backward RNN in the opposite direction (see the parallel arrows in Figure \ref{detail_model} (a)):
\begin{equation}
\begin{split}
\overrightarrow{h}_i & = \text{RNN}(\overrightarrow{h}_{i-1}, E_{w_i}) \\
\overleftarrow{h}_i & = \text{RNN}(\overleftarrow{h}_{i+1}, E_{w_i})
\end{split}
\end{equation}
where $E_{w_i} \in \mathbb{R}^{d_w}$ is the embedding for word $w_i$, and $\overrightarrow{h}_i, \overleftarrow{h}_i$ are hidden states generated in two directions. Following Bahdanau et al.~\shortcite{DBLP:journals/corr/BahdanauCB14}, we employ the Gated Recurrent Unit (GRU) as our RNN unit due to its capacity in capturing long-distance dependencies. 

We further concatenate each pair of hidden states at each time step to build a set of {\it annotation} vectors $(\mathbf{h}_1,$ $\mathbf{h}_2,$ $\ldots,$ $\mathbf{h}_{T})$,
$
\mathbf{h}_{i}^T = \left[\overrightarrow{h}_{i}^T; \overleftarrow{h}_{i}^T \right]
$.
In this way, each annotation vector $\mathbf{h}_{i}$ encodes information about the $i$-th word with respect to all the other surrounding words in the sequence. Therefore, these annotation vectors are desirable for the following modeling.

We use this encoder to represent both the source sentence $\{x_i\}_{i=1}^{T_f}$ and the target sentence $\{y_i\}_{i=1}^{T_e}$ (see the blue color in Figure \ref{detail_model}). Accordingly, our encoder generates both the source annotation vectors $\{\mathbf{h}_i\}_{i=1}^{T_f} \in \mathbb{R}^{2d_f}$ and the target annotation vectors $\{\mathbf{h}_i^{\prime}\}_{i=1}^{T_e} \in \mathbb{R}^{2d_e}$. The source vectors flow into the inferer and decoder while the target vectors the posterior approximator. 

\subsection{Variational Neural Inferer}\label{var_neu_approx}

A major challenge of variational models is how to model the latent-variable-related distributions. In VNMT, we employ neural networks to model both the prior $p_\theta(\mathbf{z}|\mathbf{x})$ and the posterior $q_\phi(\mathbf{z}|\mathbf{x}, \mathbf{y})$, and let them subject to a multivariate Gaussian distribution with a diagonal covariance structure.\footnote{The reasons of choosing Gaussian distribution are twofold: 1) it is a natural choice for modeling continuous variables; 2) it belongs to the family of ``location-scale'' distributions, which is required for the following reparameterization.} As shown in Figure \ref{graph_model}, these two distributions mainly differ in their conditions.

\subsubsection{Neural Posterior Approximator}
Exactly modeling the true posterior $p(\mathbf{z}|\mathbf{x}, \mathbf{y})$ exactly usually intractable. Therefore, 
we adopt an approximation method to simplify the posterior inference. Conventional models typically employ the {\it mean-field} approaches. However, a major limitation of this approach is its inability to capture the true posterior of $\mathbf{z}$ due to its oversimplification. Following the spirit of VAE, we use neural networks for better approximation in this paper, and assume the approximator has the following form:
\begin{equation}
q_\phi(\mathbf{z}|\mathbf{x}, \mathbf{y}) = \mathcal{N}(\mathbf{z};\mathbf{\mu}(\mathbf{x}, \mathbf{y}), \mathbf{\sigma}(\mathbf{x}, \mathbf{y})^2\mathbf{I})
\end{equation}
The mean $\mathbf{\mu}$ and s.d. $\mathbf{\sigma}$ of the approximate posterior are the outputs of neural networks based on the observed variables $\mathbf{x}$ and $\mathbf{y}$ as shown in Figure \ref{detail_model} (b). 

Starting from the variational neural encoder, we first obtain the source- and target-side representation via a {\it mean-pooling} operation over the annotation vectors, i.e.
$
\mathbf{h}_f = \frac{1}{T_f}\sum_{i}^{T_f} \mathbf{h}_{i},$$~$$ \mathbf{h}_e = \frac{1}{T_e}\sum_{i}^{T_e} \mathbf{h}_{i}^{\prime}
$. 
With these representations, we perform a non-linear transformation that projects them onto our concerned latent semantic space:
\begin{equation}\label{posterior_h_z}
\mathbf{h}_{z}^{\prime} = g(W_z^{(1)} [\mathbf{h}_f; \mathbf{h}_e] + b_{z}^{(1)})
\end{equation}
where $W_z^{(1)} \in \mathbb{R}^{d_z \times 2(d_f+d_e)}, b_{z}^{(1)} \in \mathbb{R}^{d_z}$ is the parameter matrix and bias term respectively, $d_z$ is the dimensionality of the latent space, and $g(\cdot)$ is an element-wise activation function, which we set to be $\tanh(\cdot)$ throughout our experiments.

In this latent space, we obtain the abovementioned Gaussian parameters $\mu$ and $\log \sigma^2$ through linear regression:
\begin{equation}\label{posterior_mu_sigma}
\mu = W_{\mu} \mathbf{h}_{z}^{\prime} + b_{\mu},~
\log \sigma^2 = W_{\sigma} \mathbf{h}_{z}^{\prime} + b_{\sigma}
\end{equation}
where 
$\mu,\log \sigma^2$ are both $d_z$-dimension vectors.

\subsubsection{Neural Prior Model}

Different from the posterior, we model (rather than approximate) the prior as follows:
\begin{equation}
p_\theta(\mathbf{z}|\mathbf{x}) = \mathcal{N}(\mathbf{z};\mathbf{\mu}^{\prime}(\mathbf{x}), \mathbf{\sigma}^{\prime}(\mathbf{x})^2\mathbf{I})
\end{equation}
We treat the mean $\mathbf{\mu}^{\prime}$ and s.d. $\mathbf{\sigma}^{\prime}$ of the prior as neural functions of source sentence $\mathbf{x}$ alone. This is sound and reasonable because bilingual sentences are semantically equivalent, suggesting that either $\mathbf{y}$ or $\mathbf{x}$ is capable of inferring the underlying semantics of sentence pairs, i.e., the representation of latent variable $\mathbf{z}$.

The neural model for the prior $p_\theta(\mathbf{z}|\mathbf{x})$ is the same as that (i.e. Eq (\ref{posterior_h_z}) and (\ref{posterior_mu_sigma})) for the posterior $q_\phi(\mathbf{z}|\mathbf{x}, \mathbf{y})$, except for the absence of $\mathbf{h}_e$. Besides, the parameters for the prior are independent of those for the posterior.

To obtain a representation for latent variable $\mathbf{z}$, we employ the same technique as the Eq. (\ref{vae_re_param}) and reparameterized it as 
$\mathbf{h}_z$ $=$ $\mu$ $+$ $\sigma$ $\odot$ $\epsilon$, $\epsilon$ $\sim$ $\mathcal{N}$$(0,$ $\mathbf{I})$. During decoding, however, due to the absence of target sentence $\mathbf{y}$, we set $\mathbf{h}_z$ to be the mean of $p_\theta(\mathbf{z}|\mathbf{x})$, i.e., $\mu^{\prime}$. Intuitively, the reparameterization bridges the gap between the generation model $p_\theta(\mathbf{y}|\mathbf{z},\mathbf{x})$ and the inference model $q_\phi(\mathbf{z}|\mathbf{x}, \mathbf{y})$. In other words, it connects these two neural networks. This is important since it enables the stochastic gradient optimization via standard backpropagation.

We further project the representation of latent variable $\mathbf{h}_z$ onto the target space for translation:
\begin{equation}
\mathbf{h}_e^{\prime} = g(W_z^{(2)} \mathbf{h}_z + b_z^{(2)})
\end{equation}
where $\mathbf{h}_e^{\prime} \in \mathbb{R}^{d_e^\prime}$.
The transformed $\mathbf{h}_e^{\prime}$ is then integrated into our decoder. Notice that because of the noise from $\epsilon$, the representation $\mathbf{h}_e^{\prime}$ is not fixed for the same source sentence and model parameters. This is crucial for VNMT to learn to avoid overfitting.

\subsection{Variational Neural Decoder}\label{var_neu_decoder}

Given the source sentence $\mathbf{x}$ and the latent variable $\mathbf{z}$, our decoder defines the probability over translation $\mathbf{y}$ as a joint probability of ordered conditionals:
\begin{align}
p(\mathbf{y}|\mathbf{z},\mathbf{x}) = &\prod_{j=1}^{T_e} p(y_j|y_{< j}, \mathbf{z}, \mathbf{x}) \\
\text{where}\quad p(y_j|y_{< j}, &\mathbf{z}, \mathbf{x}) = g^{\prime}(y_{j-1}, s_{j-1}, c_j) \notag
\end{align}
The feed forward model $g^{\prime}(\cdot)$ (see the yellow arrows in Figure \ref{detail_model}) and context vector $c_j = \sum_i \alpha_{ji}\mathbf{h}_i$ (see the ``$\oplus$'' in Figure \ref{detail_model}) are the same as~\cite{DBLP:journals/corr/BahdanauCB14}. The difference between our decoder and Bahdanau et al.'s decoder~\shortcite{DBLP:journals/corr/BahdanauCB14} lies in that in addition to the context vector, our decoder integrates the representation of the latent variable, i.e. $\mathbf{h}_e^{\prime}$, into the computation of $s_j$, which is denoted by the bold dashed red arrow in Figure \ref{detail_model} (c).

Formally, the hidden state $s_j$ in our decoder is calculated by\footnote{We omit the bias term for clarity.}
\begin{align*}
&s_j = (1 - u_j) \odot s_{j-1} + u_j \odot \tilde{s}_{j}, \\
&\tilde{s}_j = \tanh(W E_{y_j} + U[r_j \odot s_{j-1}] + Cc_j + V\mathbf{h}_e^{\prime}) \\
&u_j = \sigma(W_u E_{y_j} + U_u s_{j-1} + C_u c_j + V_u\mathbf{h}_e^{\prime}) \\
&r_j = \sigma(W_r E_{y_j} + U_r s_{j-1} + C_r c_j + V_r\mathbf{h}_e^{\prime})
\end{align*}
Here, $r_j$, $u_j$, $\tilde{s}_j$ denotes the reset gate, update gate and candidate activation in GRU respectively, and $E_{y_j} \in \mathbb{R}^{d_w}$ is the word embedding for target word. $W,$ $W_u,$ $W_r$ $\in$ $\mathbb{R}^{d_e \times d_w}$, $U,$ $U_u,$ $U_r$ $\in$ $\mathbb{R}^{d_e \times d_e}$, $C,$ $C_u,$ $C_r$ $\in$ $\mathbb{R}^{d_e \times 2d_f}$, and $V,$ $V_u,$ $V_r$ $\in$ $\mathbb{R}^{d_e \times d_e^{\prime}}$ are parameter weights. The initial hidden state $s_0$ is initialized in the same way as Bahdanau et al.~\shortcite{DBLP:journals/corr/BahdanauCB14} (see the arrow to $s_0$ in Figure \ref{detail_model}).

In our model, the latent variable can affect the representation of hidden state $s_j$ through the gate between $r_j$ and $u_j$. This allows our model to access the semantic information of $\mathbf{z}$ indirectly since the prediction of $y_{j+1}$ depends on $s_j$. In addition, when the model learns wrong attentions that lead to bad context vector $c_j$, the semantic representation $\mathbf{h_e}^{\prime}$ can help to guide the translation process . 

\begin{table*}[t]
\begin{center}
{ \small
\begin{tabular}{c||c|llllll}
\multicolumn{1}{c||}{\bf System} &
\multicolumn{1}{|c|}{\bf MT05 } &
\multicolumn{1}{c}{\bf MT02 } &
\multicolumn{1}{c}{\bf MT03 } &
\multicolumn{1}{c}{\bf MT04 } &
\multicolumn{1}{c}{\bf MT06 } &
\multicolumn{1}{c}{\bf MT08 } &
\multicolumn{1}{c}{\bf AVG} \\
\hline
\hline
{\it Moses} & 33.68 & 34.19 & 34.39 & 35.34 & 29.20 & 22.94 & 31.21\\
\hline
\hline
{\it GroundHog} & 31.38 & 33.32 & 32.59 & 35.05 & 29.80 & 22.82 & 30.72 \\
{\it VNMT w/o KL} & 31.40 & 33.50 & 32.92 & 34.95 & 28.74 & 22.07 & 30.44 \\
\hline
{\it VNMT} & 32.25 & {\bf 34.50}$^{++}$ & 33.78$^{++}$ & {\bf 36.72}$^{\Uparrow++}$ & {\bf 30.92}$^{\Uparrow++}$ & {\bf 24.41}$^{\uparrow++}$ & {\bf 32.07} \\

\end{tabular}
}
\end{center}
\caption{\label{performance} BLEU scores on the NIST Chinese-English translation task. {\bf AVG} = average BLEU scores on test sets. 
We highlight the best results in bold for each test set. ``$\uparrow$/$\Uparrow$'': significantly better than {\it Moses} ($p$ $<$ $0.05$/$p$ $<$ $0.01$); ``$+$/$++$'': significantly better than {\it GroundHog} ($p$ $<$ $0.05$/$p$ $<$ $0.01$);}
\end{table*}

\subsection{Model Training}

We use the Monte Carlo method to approximate the expectation over the posterior in Eq. (\ref{vnmt_vlb_org}), i.e. 
$
\mathbb{E}_{q_\phi(\mathbf{z}|\mathbf{x},\mathbf{y})}[\cdot] \simeq \frac{1}{L} \sum_{l=1}^{L} \log p_\theta(\mathbf{y}|\mathbf{x}, \mathbf{h}_{z}^{(l)})
$, 
where $L$ is the number of samples. 
The joint training objective for a training instance $(\mathbf{x}, \mathbf{y})$ is defined as follows:
\begin{align}\label{final_object}
\mathcal{L}(\theta, \phi) \simeq  -\text{KL}(q_\phi(\mathbf{z}|\mathbf{x}, \mathbf{y})||&p_\theta(\mathbf{z}|\mathbf{x}))
\notag \\
+ \frac{1}{L}\sum_{l=1}^{L} \sum_{j=1}^{T_e} \log p_\theta (y_j| &y_{< j}, \mathbf{x}, \mathbf{h}_{z}^{(l)}) \\
\text{where}~\mathbf{h}_{z}^{(l)} = \mu + \sigma \odot \epsilon^{(l)}~\text{and}&~\epsilon^{(l)}\sim \mathcal{N}(0,\mathbf{I}) \notag
\end{align}
The first term is the KL divergence between two Gaussian distributions which can be computed and differentiated without estimation (see~\cite{kingma2014autoencoding} for details). And the second term is the approximate expectation, which is also differentiable. Suppose that $L$ is 1 (which is used in our experiments), then our second term will be degenerated to the objective of conventional NMT. Intuitively, VNMT is exactly a regularized version of NMT, where the introduced noise $\epsilon$ increases its robustness, and reduces overfitting. We verify this point in our experiments.


Since the objective function in Eq. (\ref{final_object}) is differentiable, we can optimize the model parameter $\theta$ and variational parameter $\phi$ jointly using standard gradient ascent techniques. 

\section{Experiments}


\subsection{Setup}

To evaluate the effectiveness of the proposed VNMT, we conducted experiments on both Chinese-English and English-German translation tasks. Our Chinese-English training data\footnote{This corpus consists of LDC2003E14, LDC2004T07, LDC2005T06, LDC2005T10 and LDC2004T08 (Hong Kong Hansards/Laws/News).} consists of 2.9M sentence pairs, with 80.9M Chinese words and 86.4M English words respectively. We used the NIST MT05 dataset as the development set, and the NIST MT02/03/04/06/08 datasets as the test sets for the Chinese-English task. Our English-German training data\footnote{This corpus is from the WMT'14 training data~\cite{jean-EtAl:2015:ACL-IJCNLP,luong-pham-manning:2015:EMNLP}} consists of 4.5M sentence pairs with 116M English words and 110M German words\footnote{The preprocessed data can be found and downloaded from http://nlp.stanford.edu/projects/nmt/}. We used the newstest2013 (3000 sentences) as the development set, and the newstest2014 (2737 sentences) as the test set for English-German translation. We employed the case-insensitive BLEU-4~\cite{PapineniEtAl2002} metric to evaluate translation quality, and paired bootstrap sampling~\cite{koehn04} for significance test.

We compared our model against two state-of-the-art SMT and NMT systems: 
\begin{itemize}
\item {\it Moses}~\cite{Koehn:2007:MOS:1557769.1557821}: a phrase-based SMT system.
\item {\it GroundHog}~\cite{DBLP:journals/corr/BahdanauCB14}: an attention-based NMT system.
\end{itemize}
Additionally, we also compared with a variant of {\it VNMT}, which does not contain the KL part in the objective ({\it VNMT w/o KL}). This is achieved by setting $\mathbf{h}_z$ to $\mu^{\prime}$.

For {\it Moses}, we adopted all the default settings except for the language model. We trained a 4-gram language model on the Xinhua section of the English Gigaword corpus (306M words) using the SRILM\footnote{http://www.speech.sri.com/projects/srilm/download.html} toolkit with modified Kneser-Ney smoothing. Importantly, we used all words in the vocabulary.

For {\it GroundHog}, we set the maximum length of training sentences to be 50 words, and preserved the most frequent 30K (Chinese-English) and 50K (English-German) words as both the source and target vocabulary , covering approximately 98.9\%/99.2\% and 97.3\%/93.3\% on the source and target side of the two parallel corpora respectively
. All other words were represented by a specific token ``UNK''. Following Bahdanau et al.~\shortcite{DBLP:journals/corr/BahdanauCB14}, we set $d_w$ $=$ $620$, $d_f$ $=$  $1000$, $d_e$ $=$ $1000$, and $M$ $=$ $80$. All other settings are the same as the default configuration (for {\it RNNSearch}). During decoding, we used the beam-search algorithm, and set beam size to 10.

\begin{table*}[t]
\begin{center}
{ \small
\begin{tabular}{c||c|llllll}
\multicolumn{1}{c||}{\bf System} &
\multicolumn{1}{|c|}{\bf MT05 } &
\multicolumn{1}{c}{\bf MT02 } &
\multicolumn{1}{c}{\bf MT03 } &
\multicolumn{1}{c}{\bf MT04 } &
\multicolumn{1}{c}{\bf MT06 } &
\multicolumn{1}{c}{\bf MT08 }\\
\hline
\hline
{\it GroundHog} & 18.23 & 22.20 & 20.19 & 21.67 & 19.11 & 13.41\\
\hline
{\it VNMT} & {\bf 21.31} & {\bf 26.02} & {\bf 23.78} & {\bf 25.81} & {\bf 21.81} & {\bf 15.59} \\
\end{tabular}
}
\end{center}
\caption{\label{performance_newdata} BLEU scores on the new dataset. All improvements are significant at $p$ $<$ $0.01$.}
\end{table*}

\begin{table*}[t]
\begin{center}
{ \small
\begin{tabular}{l|l|l}
\multicolumn{1}{l|}{\bf System} &
\multicolumn{1}{l|}{\bf Architecture } &
\multicolumn{1}{c}{\bf BLEU }\\
\hline
\hline
\multicolumn{3}{c}{\it Existing end-to-end NMT systems} \\
\hline
Jean et al.~\shortcite{jean-EtAl:2015:ACL-IJCNLP} & RNNSearch & 16.46 \\
Jean et al.~\shortcite{jean-EtAl:2015:ACL-IJCNLP} & RNNSearch + unk replace & 18.97 \\
Jean et al.~\shortcite{jean-EtAl:2015:ACL-IJCNLP} & RNNsearch + unk replace + large vocab & 19.40 \\
Luong et al.~\shortcite{luong-pham-manning:2015:EMNLP} & LSTM with 4 layers + dropout + local att. + unk replace & 20.90 \\
\hline
\multicolumn{3}{c}{\it Our end-to-end NMT systems} \\
\hline
\multirow{3}{*}{\it this work} & RNNSearch & 16.40 \\
& VNMT & 17.13$^{\scriptstyle ++}$ \\
& VNMT + unk replace & 19.58$^{\scriptstyle ++}$ \\
\end{tabular}
}
\end{center}
\caption{\label{english_german_translation} BLEU scores on the English-German translation task.}
\end{table*}

For {\it VNMT}, we initialized its parameters with the trained {\it RNNSearch} model. The settings of our model are the same as that of {\it GroundHog}, except for some parameters specific to VNMT. Following VAE, we set the sampling number $L$ $=$ $1$. Additionally, we set $d_e^{\prime}$ = $d_z$ $=$ $2d_f$ $=$ $2000$ according to preliminary experiments. We used the Adadelta algorithm for model training with $\rho=0.95$. With regard to the source and target encoders, we shared their recurrent parameters but not word embeddings. 

We implemented our VNMT based on {\it GroundHog}\footnote{Our code is publicly available at https://github.com/DeepLearnXMU/VNMT.}. Both NMT systems are trained on a Telsa K40 GPU. In one hour, {\it GroundHog} processes about 1100 batches, while our {\it VNMT} processes 630 batches. 

\subsection{Results on Chinese-English Translation}

Table \ref{performance} summarizes the BLEU scores of different systems on the Chinese-English translation tasks. 
Clearly VNMT significantly improves translation quality in terms of BLEU on most cases, and obtains the best average results that gain 0.86 and 1.35 BLEU points over {\it Moses} and {\it GroundHog} respectively. Besides, without the KL objective, {\it VNMT w/o KL} obtains even worse results than GroundHog. These results indicate the following two points: 1) explicitly modeling underlying semantics by a latent variable indeed benefits neural machine translation, and 2) the improvements of our model are not from enlarging the network.

\begin{figure}[t]
\centering
\includegraphics[scale=0.70]{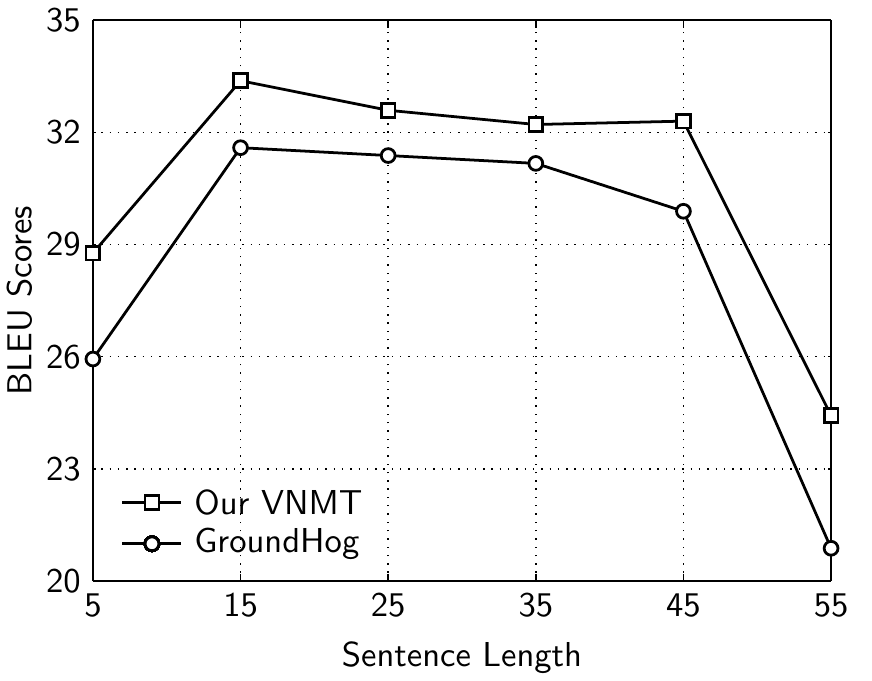}
\caption{\label{length_work} BLEU scores on different groups of source sentences in terms of their length.}
\end{figure}

\begin{table*}[t]
\begin{center}
{ \small
\begin{tabular}{c||p{1.4\columnwidth}}
\hline
\multirow{4}{*}{\it Source} & \chinese{\textcolor{zb_red}{两~国~官员}~确定~了~今后~会谈~的~日程~和~模式~,~建立~起~进行~持续~对话~的~机制~,~此举~标志~着~巴~印~对话~进程~在~中断~两~年~后~重新~启动~,~为~两~国~逐步~解决~包括~克什米~尔~争端~在内~的~所有~悬而未决~的~问题~奠定~了~基础~,~\textcolor{zb_red}{体现~了~双方~可贵~的~和平~诚意~。}} \\
\hline
\multirow{6}{*}{\it Reference} & {\it \textcolor{zb_red}{the officials of the two countries} have established the mechanism for continued dialogue down the road, including a confirmed schedule and model of the talks.  this symbolizes the restart of the dialogue process between pakistan and india after an interruption of two years and has paved a foundation for the two countries to sort out gradually all the questions hanging in the air, including the kashmir dispute.  \textcolor{zb_red}{it is also a realization of their precious sincerity for peace.}} \\
\hline
\multirow{5}{*}{\it Moses} & {\it officials of the two countries set the agenda for future talks , and the pattern of a continuing dialogue mechanism . this marks a break in the process of dialogue between pakistan and india , two years after the restart of the two countries including kashmir dispute to gradually solve all the outstanding issues have laid the foundation of the two sides showed great sincerity in peace .} \\
\hline
\multirow{4}{*}{\it GroundHog} & {\it \textcolor{zb_red}{the two countries} have decided to set up a mechanism for conducting continuous dialogue on the agenda and mode of the talks . this indicates that the ongoing dialogue between the two countries has laid the foundation for the gradual settlement of all outstanding issues including the dispute over kashmir .}\\
\hline
\multirow{5}{*}{\it VNMT} & {\it \textcolor{zb_red}{the officials of the two countries} set up a mechanism for holding a continuous dialogue on the agenda and mode of the future talks, and this indicates that the ongoing dialogue between pakistan and india has laid a foundation for resolving all outstanding issues , including the kashmir disputes , \textcolor{zb_red}{and this serves as a valuable and sincere peace sincerity .} }\\
\hline
\end{tabular}
}
\end{center}
\caption{\label{sentence_analysis} Translation examples of different systems. We highlight important parts in red color.}
\end{table*}

\subsection{Results on Long Sentences}

We further testify VNMT on long sentence translation where the vanilla NMT usually suffers from attention failures~\cite{DBLP:journals/corr/TuLLLL16,2016arXiv160804631B}. We believe that the global latent variable can play an important role on long sentence translation.

Our first experiment is carried out on 
6 disjoint groups according to the length of source sentences in our test sets. Figure \ref{length_work} shows the BLEU scores of two neural models. We find that the performance curve of our VNMT model always appears to be on top of that of {\it GroundHog} with a certain margin. Specifically, on the final group with the longest source sentences, our VNMT obtains the biggest improvement (3.55 BLEU points). Overall, these obvious improvements on all groups in terms of the length of source sentences indicate that the global guidance from the latent variable benefits our VNMT model. 

Our second experiment is carried out on a synthetic dataset where each new source sentence is a concatenation of neighboring source sentences in the original test sets. As a result, the average length of source sentences in the new dataset ($>$ 50) is almost twice longer than the original one.  Translation results is summarized in Table \ref{performance_newdata}, where our VNMT obtains significant improvements on all new test sets. This further demonstrates the advantage of introducing the latent variable.


\subsection{Results on English-German Translation}

Table \ref{english_german_translation} shows the results on English-German translation. We also provide several existing NMT systems that use the same training, development and testing data. The results show that VNMT significantly outperforms GroundHog and achieves a significant gain of 0.73 BLEU points ($p < 0.01$). 
With unknown word replacement~\cite{jean-EtAl:2015:ACL-IJCNLP,luong-pham-manning:2015:EMNLP}, VNMT reaches the performance level that is  comparable to the previous state-of-the-art NMT results. 


\subsection{Translation Analysis}

Table \ref{sentence_analysis} shows a translation example that helps understand the advantage of VNMT over NMT
. As the source sentence in this example is long (more than 40 words), the translation generated by {\it Moses} is relatively messy and incomprehensible. In contrast, translations generated by neural models (both {\it GroundHog} and {\it VNMT}) are much more fluent and comprehensible. However, there are essential differences between {\it GroundHog} and our {\it VNMT}. Specifically, {\it GroundHog} does not translate the phrase ``\chinese{官员}'' at the beginning of the source sentence. The translation of the clause ``\chinese{体现~了~双方~可贵~的~和平~诚意~。}'' at the end of the source sentence is completely lost. In contrast, our VNMT model does not miss or mistake these fragments and can convey the meaning of entire source sentence to the target side.

From these examples, we can find that although attention networks can help NMT trace back to relevant parts of source sentences for predicting target translations, capturing the semantics of entire sentences still remains a big challenge for neural machine translation. Since NMT implicitly models variable-length source sentences with fixed-size hidden vectors, some details of source sentences (e.g., the red sequence of words in Table \ref{sentence_analysis}) may not be encoded in these vectors at all. VNMT seems to be able to capture these details through a latent variable that explicitly model underlying semantics of source sentences. The promising results suggest that VNMT provides a new mechanism to deal with sentence semantics.

\section{Related Work}


\subsection{Neural Machine Translation}


Neural machine translation starts from the sequence to sequence learning, where Sutskever et al.~\shortcite{DBLP:journals/corr/SutskeverVL14} employ two multilayered Long Short-Term Memory (LSTM) models that first encode a source sentence into a single vector and then decode the translation word by word until a special end token  
is generated. In order to deal with issues caused by encoding all source-side information into a fixed-length vector, Bahdanau et al.~\shortcite{DBLP:journals/corr/BahdanauCB14} introduce attention-based NMT that aims at automatically concentrating on relevant source parts for predicting target words during decoding. The incorporation of attention mechanism allows NMT to cope better with long sentences, and makes it really comparable to or even superior to conventional SMT.

Following the success of attentional NMT, a number of approaches and models have been proposed for NMT recently, which can be grouped into different categories according to their motivations: dealing with rare words or large vocabulary~\cite{jean-EtAl:2015:ACL-IJCNLP,luong-EtAl:2015:ACL-IJCNLP,2015arXiv150807909S}, learning better attentional structures~\cite{luong-pham-manning:2015:EMNLP}, integrating SMT techniques~\cite{2015arXiv151204650C,2015arXiv151202433S,2016arXiv160103317F,DBLP:journals/corr/TuLLLL16}, memory network~\cite{2015arXiv150606442M}, etc. All these models are designed within the discriminative encoder-decoder framework, leaving the explicit exploration of underlying semantics with a variational model an open problem.

\subsection{Variational Neural Model}

In order to perform efficient inference and learning in directed probabilistic models on large-scale dataset, Kingma and Welling~\shortcite{kingma2014autoencoding} as well as Rezende et al.~\shortcite{DBLP:conf/icml/RezendeMW14} introduce variational neural networks. Typically, these models utilize an neural inference model to approximate the intractable posterior, and optimize model parameters jointly with a reparameterized variational lower bound using the standard stochastic gradient technique. This approach is of growing interest due to its success in various tasks.

Kingma et al.~\shortcite{DBLP:conf/nips/KingmaMRW14} revisit the approach to semi-supervised learning with generative models and further develop new models that allow effective generalization from a small labeled dataset to a large unlabeled dataset. Chung et al.~\shortcite{DBLP:journals/corr/ChungKDGCB15} incorporate latent variables into the hidden state of a recurrent neural network, while Gregor et al.~\shortcite{DBLP:journals/corr/GregorDGW15} combine a novel spatial attention mechanism that mimics the foveation of human eyes, with a sequential variational auto-encoding framework that allows the iterative construction of complex images. Very recently, Miao et al.~\shortcite{2015arXiv151106038M} propose a generic variational inference framework for generative and conditional models of text.

The most related work is that of Bowman et al.~\shortcite{2015arXiv151106349B}, where they develop a variational autoencoder for unsupervised generative language modeling. The major difference is that they focus on the monolingual language model, while we adapt this technique to bilingual translation. Although variational neural models have been widely used in NLP tasks and the variational decoding has been investigated for SMT~\cite{li-eisner-khudanpur:2009:ACLIJCNLP}, the adaptation and utilization of variational neural model to neural machine translation, to the best of our knowledge, has never been investigated before.

\section{Conclusion and Future Work}

In this paper, we have presented a variational model for neural machine translation that incorporates a continuous latent variable to model the underlying semantics of sentence pairs. We approximate the posterior distribution with neural networks and reparameterize the variational lower bound. This enables our model to be an end-to-end neural network that can be optimized through the stochastic gradient algorithms. Comparing with the conventional attention-based NMT, our model is better at translating long sentences. It also greatly benefits from a special regularization term brought with this latent variable. Experiments on Chinese-English and English-German translation tasks verified the effectiveness of our model.

In the future, since the latent variable in our model is at the sentence level, we want to explore more fine-grained latent variables for neural machine translation, such as the {\it Recurrent Latent Variable Model}~\cite{DBLP:journals/corr/ChungKDGCB15}. We are also interested in applying our model to other similar tasks. 

\section*{Acknowledgments}

The authors were supported by National Natural Science Foundation of China (Grant Nos 61303082, 61672440, 61622209 and 61403269), Natural Science Foundation of Fujian Province (Grant No. 2016J05161), Natural Science Foundation of Jiangsu Province (Grant No. BK20140355), and Research fund of the Provincial Key Laboratory for Computer Information Processing Technology in Soochow University (Grant No. KJS1520).
We also thank the anonymous reviewers for their insightful comments.

\bibliography{emnlp2016}
\bibliographystyle{emnlp2016}

\end{document}